\documentclass{ifacconf}
\usepackage{graphicx}      
\usepackage{natbib}        
\usepackage{amsmath,amssymb,amsfonts,bm,mathtools}
\usepackage{comment}
\usepackage{siunitx}
\usepackage{xcolor}
\usepackage{mathtools}
\usepackage{fp, tikz, pgfplots}

\usepackage[normalem]{ulem}
\usepackage{algorithmic}
\usepackage{algorithm}
\usepackage{subfigure}
\usepackage{multirow}
\usepackage{booktabs} 
\usepackage{tabularx}
\usepackage{comment}

\usepackage{pgfplotstable}

\usepackage{array}
\newcolumntype{V}{>{\setbox0=\hbox\bgroup}c<{\egroup}@{}}

\usetikzlibrary{arrows,shapes,backgrounds,patterns,fadings,matrix,arrows,calc,
	intersections,decorations.markings,
	positioning,external,arrows.meta}
\tikzset{
	block/.style = {draw, rectangle,
		minimum height=1cm,
		minimum width=2cm},
	input/.style = {coordinate,node distance=1cm},
	output/.style = {coordinate,node distance=6cm},
	arrow/.style={draw, -latex,node distance=2cm},
	pinstyle/.style = {pin edge={latex-, black,node distance=2cm}},
	sum/.style = {draw, circle, node distance=1cm},
}
\usepgfplotslibrary{fillbetween}
\definecolor{nicegreen}{RGB}{0,150,0}

\def\loadsomeimgs{1}
\def\tblincmse{0}

\makeatletter
\renewcommand*{\@fnsymbol}[1]{\ensuremath{\ifcase#1\or * \or ** \or *** \or \dagger \or
		\mathsection\or \mathparagraph\or \|\or **\or \dagger\dagger
		\or \ddagger\ddagger \else\@ctrerr\fi}}
\makeatother

\pgfplotsset{table/col sep = comma}

\def\1{\bm{1}}

\def\ri{{\textnormal{i}}}

\def\vf{{\bm{f}}}
\def\vg{{\bm{g}}}

\def\vu{{\bm{u}}}

\def\vx{{\bm{x}}}
\def\vy{{\bm{y}}}

\DeclareMathAlphabet{\mathsfit}{\encodingdefault}{\sfdefault}{m}{sl}
\SetMathAlphabet{\mathsfit}{bold}{\encodingdefault}{\sfdefault}{bx}{n}

\def\sD{{\mathbb{D}}}

\def\sN{{\mathbb{N}}}

\def\sR{{\mathbb{R}}}

\def\sU{{\mathbb{U}}}

\def\sX{{\mathbb{X}}}
\def\sY{{\mathbb{Y}}}

\newcommand{\R}{\mathbb{R}}

\DeclareMathOperator*{\argmin}{arg\,min}

\def\Nepi{{N_\text{epi}}}
\def\Ntr{{N_\text{tr}}}
\def\Nte{{N_\text{te}}}
\def\dx{{d_x}}
\def\dy{{d_y}}

\def\du{{d_u}}
\def\dimp{{d_p}}
\def\yepi{{\tilde{y}}}
\def\xepi{{\tilde{\vx}}}

\def\vxtr{{\vx}}
\def\xtr{{x}}

\def\epi{\eta}

\def\acp{{\alpha}}

\def\sXtr{\sX_{\text{tr}}}

\def\sYtr{\sY_{\text{tr}}}
\def\sDtr{\sD_{\text{tr}}}
\def\sRzp{\sR_{0,+}}

\def\sXepi{\sX_{\text{epi}}}
\def\siXepi{\hat{\sX}_{\text{epi}}}
\def\sYepi{\sY_{\text{epi}}}
\def\sDepi{\sD_{\text{epi}}}

\def\distU{\mathcal{U}}
\def\distN{\mathcal{N}}

\def\distB{\mathcal{B}}

\def\bigO{{\mathcal{O}}}

\def\EpiOut{\textit{EpiOut}}
\def\Dropout{\textit{Dropout}}
\def\BNN{\textit{BNN}}
\def\diag{\text{diag}}

\newcommand{\new}[1]{{\color{black}#1}}

\begin{document}
\begin{frontmatter}

\title{
 Deep Learning based Uncertainty Decomposition for Real-time Control 
} 


\thanks[footnoteinfo]{\new{This work was supported by the European Research Council (ERC) Consolidator Grant "Safe data-driven control for human-centric systems (CO-MAN)" under grant agreement number 864686, and TUM AGENDA 2030, funded by the Federal Ministry of Education and Research (BMBF) and the Free State of Bavaria under the Excellence Strategy of the Federal Government and the Länder as well as by the Hightech Agenda Bavaria.}}

\author[First]{Neha Das$^\dagger$} 
\author[First]{Jonas Umlauft$^\dagger$} 
\author[First]{Armin Lederer} 
\author[First]{Alexandre Capone} 
\author[Second]{Thomas Beckers} 
\author[First]{Sandra Hirche} 

\address[First]{
School of Computation, Information and Technology, Technical University of Munich, Munich, Germany (e-mail: \{neha.das, jonas.umlauft, armin.lederer, alexandre.capone, hirche\}@tum.de)}
\address[Second]{Department of Computer Science, Vanderbilt University, Nashville, TN 37212, USA (e-mail: thomas.beckers@vanderbilt.edu)}

\begin{abstract}                
Data-driven control in unknown environments requires a clear understanding of the involved uncertainties for ensuring safety and efficient exploration.
While aleatoric uncertainty that arises from measurement noise can often be explicitly modeled given a parametric description, it can be harder to model epistemic uncertainty, which describes the presence or absence of training data. The latter can be particularly useful for implementing exploratory control strategies when system dynamics are unknown. We propose a novel method for detecting the absence of training data using deep learning, which gives a continuous valued scalar output between $0$ (indicating low uncertainty) and $1$ (indicating high uncertainty). We utilize this detector as a proxy for epistemic uncertainty and show its advantages over existing approaches on synthetic and real-world datasets. Our approach can be directly combined with aleatoric uncertainty estimates and allows for uncertainty estimation in real-time as the inference is sample-free unlike existing approaches for uncertainty modeling. We further demonstrate the practicality of this uncertainty estimate in deploying online data-efficient control on a simulated quadcopter acted upon by an unknown disturbance model.
\end{abstract}


\end{frontmatter}

\renewcommand*{\thefootnote}{\fnsymbol{footnote}}
\footnotetext[4]{Both authors contributed equally.}
\footnotetext{©2023 the authors. This work has been accepted to IFAC for publication under a Creative Commons Licence CCBY-NC-ND.}
\renewcommand*{\thefootnote}{\arabic{footnote}}

\section{Introduction}\label{sec:intro}

Data-driven approaches are increasingly applied for system identification in safety-critical domains such as autonomous driving or human-robot interaction~\citep{grigorescu2020survey}. These techniques typically infer the unknown system dynamics from a dataset of measurements and use the inferred dynamics model for control. However, these datasets often suffer from noise and sparsity which can lead to modeling errors. 
Many control strategies have benefited from estimating the model's fidelity in addition to the point estimate of the system outputs. In particular, estimates of the model's fidelity have been incorporated in risk-averse control strategies 
to mitigate the risks induced by modeling errors \citep{umlauft2018uncertainty}, as well as for safe exploration 
\citep{liu2020robust}, event-triggered learning \citep{umlauft2020feedback}, data-efficient exploration \citep{capone2020data} among others.

Many of these approaches differentiate between the exploitation of the two different sources of the uncertainty - measurement noise, which is irreducible, also commonly referred to as aleatoric uncertainty, and epistemic uncertainty, that arises from data scarcity and can be combated by incorporating more measurements \citep{gal2016uncertainty}.  For example, epistemic uncertainty is utilized by  \cite{umlauft2020feedback} for event-triggered learning 
and \cite{depeweg2018decomposition} lays down a framework for risk-sensitive online learning which can explicitly balance between the aleatoric and epistemic uncertainty components.  
Implementation of such approaches necessitates the decomposition of epistemic and aleatoric uncertainty in real-time. In particular, our interest lies in the implementation of a uncertainty-based event-triggered adaptive control approach for mobile platforms with large (nearly \SI{1}{\kilo\hertz}) sampling rates. The underlying idea is that with real-time uncertainty decomposition, epistemic uncertainty estimates can be obtained at or close to the system's sampling rate and measurements taken for learning the dynamics model can be limited to regions with high epistemic uncertainty, thus increasing data-efficiency while at the same time providing a means to adapt control for better tracking performance.

The field of uncertainty-estimation encompasses a wide-variety of approaches. 
Gaussian processes (GPs) inherently provide a measure for their own fidelity with  posterior standard deviation \citep{rasmussen2006gaussian} 
and can differentiate between the two kinds of uncertainties but generally suffer from poor scaling to large datasets \citep{quinonero2005unifying}. 
While various methods have been proposed to make GP computationally more efficient~\citep{snelson2005sparse, titsias2009variational} these introduce approximations that can distort the uncertainty prediction 
\citep{liu2020gaussian}.

More recently, several different approaches for uncertainty decomposition using deep learning frameworks have been proposed. Popular approaches rely on Bayesian approximations \citep{depeweg2016learning} or permanent dropouts \citep{gal2016uncertainty}. Furthermore, latent inputs can also be  used to achieve a decomposition into aleatoric and epistemic uncertainty as presented by~\citet{depeweg2018decomposition}. 
However, in particular for Bayesian NNs, these approaches become computationally challenging
~\citep{kwon2020uncertainty}. Furthermore, the prediction requires sampling the entire network before the statistics of the output can be computed. When applied to control problems that are real-time critical e.g., robotics with a sampling rate up to \SI{1}{\kilo\hertz}, these computational burdens prohibit an employment of these techniques. 
Ensemble-based approaches \citep{lakshminarayanan2017simple}, that utilize an ensemble of individual models can also provide epistemic uncertainty estimates, but are also sample based and have significant memory requirements.
A sampling-free estimation method is proposed by \citet{postels2019sampling}, but suffers from a quadratic space complexity in the number of weights in the network and relies on first-order Taylor approximations in the propagation of the uncertainties, which can become inaccurate depending on the non-linearity of the activation functions. 

In contrast to the above approaches, we propose a deep learning framework that is able to estimate both epistemic and aleatoric uncertainty in real-time along with the model prediction without adding to the storage requirements of the model beyond that of an additional layer of neurons. Our method models aleatoric uncertainty by extending the neural network that models the dynamics with an additional head and incorporating the output into the training loss function similar to \citet{kendall2017what}. For estimating epistemic uncertainty, our method learns a classifier that can detect the absence of training data or in other words, detect whether a test sample is out-of-distribution (OOD). We evaluate the proposed method on low-dimensional synthetic and real-world benchmark datasets both qualitatively and quantitatively. 
To demonstrate the applicability of real-time uncertainty decomposition via our method, we employ the proposed epistemic uncertainty estimation method to learn a data-efficient controller for a simulated quadcopter navigating a space with unknown disturbances online while tuning controller gains in accordance with the aleatoric uncertainty estimates that our model outputs. The different treatment of the epistemic and aleatoric uncertainty measures is in contrast to previous works 
that take a general uncertainty measure for risk-averse control.

The remainder of this paper is structured as follows. In Sec.~\ref{sec:problem}, we formalize our problem setting. Our approach for uncertainty decomposition for real-time control is described in Sec.~\ref{sec:online_control}. We evaluate our approach in Sec.~\ref{sec:Eval} before concluding the paper in Sec.~\ref{sec:conc}. 

\section{Problem formulation}
\label{sec:problem}
We consider the discrete-time dynamical system with control~$\vu \in \sU \subseteq \R^{\du}$ and state~$\vx \in \sX \subseteq \R^{\dx}$
\begin{align}\label{eq:sys}
	\vx_{k+1} = \vg(\vx_k,\vu_k) + \vy_k,
\end{align}
where~$\vg\colon \sX \times \sU \to \sX$ is the known part of the system, while $\vy \in \sY \subseteq \R^{\dy}$ represents a disturbance composed of the unmodeled part $\vf$ of the dynamics and a Gaussian distributed noise $\epsilon$:
\begin{align}\label{eq:true}
	\vy_k = \vf(\vx_k) + \epsilon, \quad \epsilon \in \mathcal{N}(0, \diag(\bm{\sigma}(\vx))),
\end{align} 
where the diagonal noise covariance $\bm{\sigma} \in \R^{\dy}$ is unknown and cannot be measured independently of $\vy$. While $\vf$ and $\bm{\sigma}$ are shown to be a function of $\vx$ only, it is straightforward to extend them to be a function of $\vu$ as well. However, we omit this for the sake of brevity.

Given that measurements can be taken to obtain the data set $\sDtr = \{(\vx_i,\vy_i)\}_{i=1}^\Ntr$ with inputs~$\sXtr = \{\vx_i\}_{i=1}^\Ntr$ and outputs~$\sYtr = \{\vy_i\}_{i=1}^\Ntr$, our goal is to learn an uncertainty aware model of the stochastic process underlying $\vy$.

Uncertainty in data-driven models arises from two distinct sources and is thus categorized into two types. The uncertainty in data, known as \textit{aleatoric} uncertainty is inherent to the stochastic process underlying $\vy$ and is irreducible. However, with the knowledge that the disturbance is Gaussian distributed, one can model the parameters (mean and variance) of this distribution with a neural network with two outputs $[\hat\vf, \bm{\hat\sigma}]$ \citep{kendall2017what}. The other type of uncertainty, termed \textit{epistemic}, arises from the limited expressive power of the model or  absence of sufficient data for training it~\citep{gal2016uncertainty}.

In addition to modeling the true mapping $\vf$ with the neural network output $\hat \vf:\sX\to\sY$ and associated aleatoric uncertainty with $\bm{\hat{\sigma}}:\sX\to\R^{\dy}_+$, our goal includes explicitly modeling the epistemic uncertainty of our learned model $\hat \vf$ such that it can be estimated in real-time. More specifically, we assume $\eta$ to be a mapping on $\sX$ that acts as an indicator of the model's epistemic uncertainty at any input $\vx \in \sX$.

\section{\EpiOut~- Direct Estimation of Epistemic Uncertainty}\label{sec:epi}



Note that epistemic uncertainty at a $\vx \in \sX$ expresses the model $\hat \vf$'s ignorance regarding the true mapping $\vf$ emanating from the absence of training data at (or close to) $\vx$. We therefore posit that an OOD detector, which can indicate whether a sample is in training data distribution or not, can be used as a proxy for epistemic uncertainty. 

\begin{figure}
	\centering 
	\subfigure[\small \EpiOut~architecture]{
	    \resizebox{0.9\columnwidth}{!}{
                \def\layersep{1.5cm}

\begin{tikzpicture}[shorten >=1pt,->,draw=black!50, node 
distance=\layersep, scale=1.0]
\def\ni{1}
\def\nh{5}
\def\np{3}
\def\nph{3}
\def\noh{2}
\tikzstyle{every pin edge}=[<-,shorten <=1pt]
\tikzstyle{neuron}=[circle,fill=black!25,minimum size=17pt,inner 
sep=0pt]
\tikzstyle{input neuron}=[neuron, fill=green!50];
\tikzstyle{output neuron}=[neuron, fill=red!50];
\tikzstyle{parameter neuron}=[neuron, fill=yellow!80!black];
\tikzstyle{hidden neuron}=[neuron, fill=blue!70];
\tikzstyle{param hidden neuron}=[neuron, fill=yellow!80!black];
\tikzstyle{epi neuron}=[neuron, fill=red!50];
\tikzstyle{annot} = [text width=6em, text centered, font = 
\scriptsize]


\node[input neuron] (I) at (0,-2.5cm) {$x$}; 

\foreach \name / \y in {1,...,\nh}
\path[yshift=0.5cm]
node[hidden neuron] (H-\name) at (\layersep,-\y cm) {};

\foreach \name / \y in {1,...,\nph}
\path[yshift=0.5cm]
node[parameter neuron] (PH-\name) at (2*\layersep,-\y cm) {};

\foreach \name / \y in {1,...,\noh}
\path[yshift=0.5cm]
node[epi neuron] (OH-\name) at (2*\layersep,-105-\y cm) {};

\path[yshift=0.5cm]
node[parameter neuron] (P-1) at (3*\layersep,-1.5 cm) {\scriptsize$\ \hat{f}(x)\ $};
\path[yshift=0.5cm]
node[parameter neuron] (P-2) at (3*\layersep,-3 cm) {\scriptsize$\hat{\sigma}(x)^2$};
\path[yshift=0.5cm]
node[epi neuron] (P-3) at (3*\layersep,-4.5 cm) {\scriptsize$\ \epi(x)\ $};
\node[parameter neuron] (O-1) at (3.5*\layersep,-1.8 cm) {$y$};

\foreach \source in {1,...,\ni}
\foreach \dest in {1,...,\nh}
\path (I) edge (H-\dest);

\foreach \source in {1,...,\nh}
\foreach \dest in {1,...,\nph}
\path (H-\source) edge (PH-\dest);

\foreach \source in {1,...,\nh}
\foreach \dest in {1,...,\noh}
\path (H-\source) edge (OH-\dest);

\foreach \source in {1,...,\nph}
\foreach \dest in {1,...,2}
\path (PH-\source) edge (P-\dest);

\foreach \source in {1,...,\noh}
\path (OH-\source) edge (P-3);

\foreach \source in {1,...,2}
\path (P-\source) edge (O-1);


\node[annot,above of=H-1, node distance=0.7cm] (hl) {feature \\ 
	layer(s)};
\node[annot,left of=hl] {input\\ layer};
\node[annot,right of=hl, node distance=2.3cm] (phl) {o/p \& aleatoric \\ layer(s)};
\node[annot,below of=phl, node distance=3.55cm] (ohl) {epistemic \\ layer(s)};

%

\draw[->,gray] (P-3.east)-- ([xshift = 0.5cm]P-3.east) 
node[anchor=south, pos = 0.5, gray] 	{{\scriptsize fit}}
node[anchor=west, pos = 1,gray] 	{{\scriptsize cross entropy}};

\draw[->,gray] (O-1.east)-- ([xshift = 0.5cm]O-1.east) 
node[anchor=south, pos = 0.5, gray] 	{{\scriptsize fit}}
node[anchor=west,pos=1, gray] {{\scriptsize $\distN$ likelihood}};

\draw[->] (P-3.south)|- ([shift = ({0.8cm,-0.2cm})]P-3.south) 
node[anchor=west, pos = 1] {{\scriptsize epistemic uncertainty}};

\draw[->] (P-1.north)|- ([shift = ({0.8cm,0.2cm})]P-1.north) 
node[anchor=west, pos = 1] {{\scriptsize mean disturbance}};	

\draw[->] (P-2.south)|- ([shift = ({0.8cm,-0.2cm})]P-2.south) 
node[anchor=west, pos = 1] {{\scriptsize aleatoric uncertainty}};	
\end{tikzpicture}
            }
	    } 
	\hfill
	\subfigure[\small $\sXepi$ generation]{\includegraphics[width=0.45\textwidth, trim={2cm 2cm 2cm 2cm},clip]{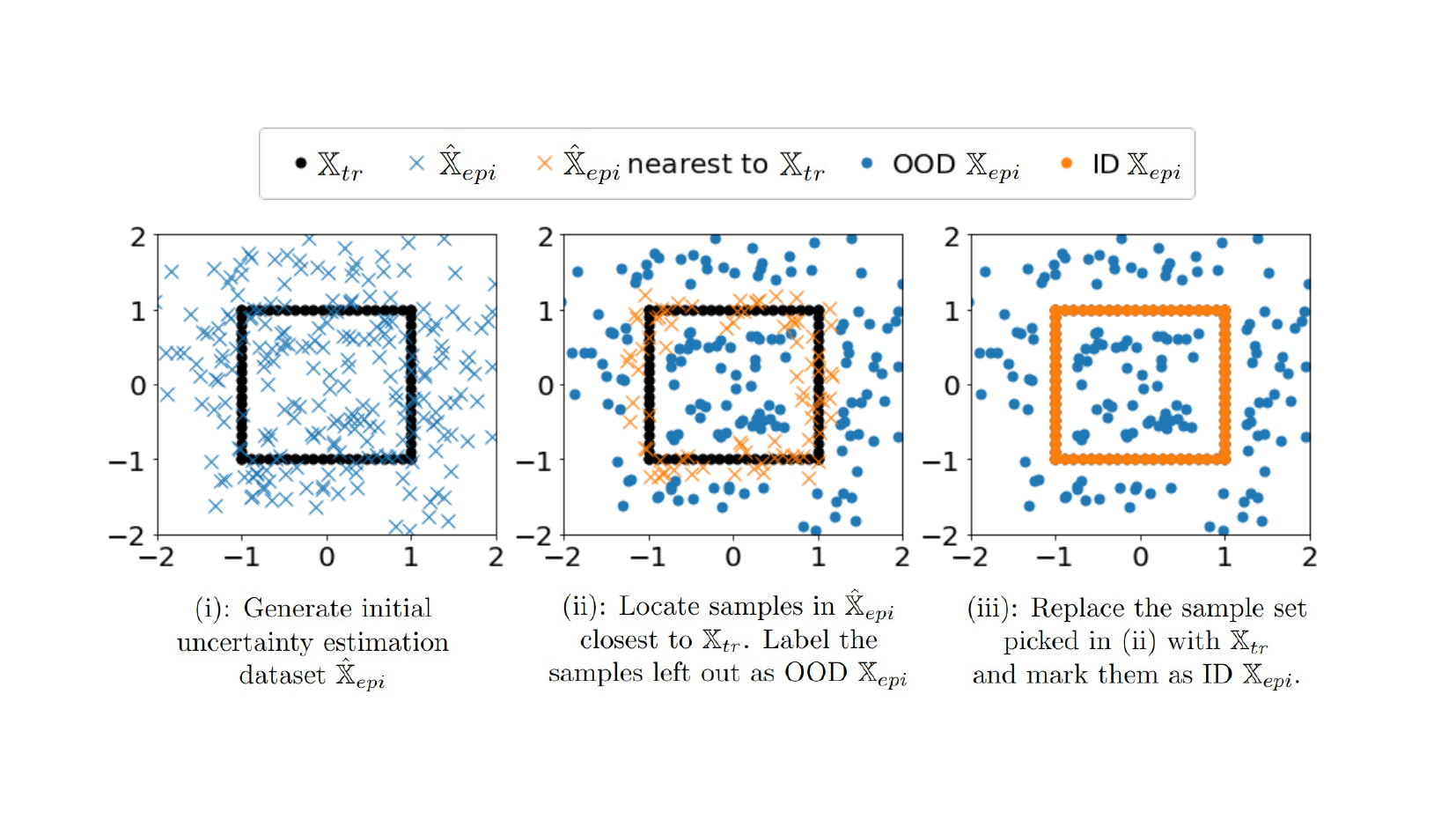}}	
	\caption{Modeling and training~\EpiOut:  (a) Schematic of an example Neural Network architecture with decomposed uncertainties (b) Process of generating samples $\mathbb{X}_{epi}$ for learning the OOD classifier.} 
	\label{fig:NN}
\end{figure}

We explicitly model this OOD detector via  a separate module consisting of an output node and optionally distinct layers termed  \EpiOut~added to the network with outputs $\hat \vf$ and $\bm{\hat{\sigma}}$ (refer to Fig.~\ref{fig:NN} for an example schematic).   
Consequently, the dataset for training \EpiOut~must be constructed to reflect this absence of data. However, we do not use a held out dataset for learning this classifier. Instead, we generate OOD samples using a simple mechanism based on the hypothesis that OOD samples near the boundary of the training distribution are sufficient for learning an OOD detector that is able to generalize to the entire OOD space. Previous works by \citet{lee2018training, marson2021boundary} have backed up this hypothesis for classification problems. Similar to our work, \citet{hafner2020noise} also proposes to generate OOD samples near the boundary of the training distribution for uncertainty estimation during regression. However, it simply adds probabilistic noise to the input to generate such samples which can result in overlap between the samples assigned OOD and training datapoints. In contrast, we follow a multi-step process that ensures the separation of the training distribution from samples assigned OOD (See Fig. \ref{fig:NN} (b)). 

In brief, (i) we first sample an initial set of points $\siXepi$ close to the boundary of training samples, (ii) then locate the $N_{tr}$ samples closest to the data and finally, (iii) replace this sample with $\sXtr$.
We therefore generate a set $\sXepi$ that contains both inputs in $\sXtr$ labeled as low uncertainty data points as well as inputs in $\sX \backslash \sXtr$, labeled as high uncertainty data points. 
We denote this epistemic dataset by $\sDepi =\{(\xepi_j,\yepi_j)\}_{j=1}^\Nepi$,  $\Nepi \in \sN$, with inputs $\sXepi = \{\xepi_j\}_{j=1}^\Nepi$ and outputs $\sYepi = \{\yepi_j\}_{j=1}^\Nepi$.

Different variations for sampling the initial set~$\siXepi$ can be chosen depending on the desired scalability properties. A naive approach is to sample the entire input space uniformly, which suffers from the curse of dimensionality. 

Alternatively, we propose to sample \new{$n \in \mathbb{N}$} 
points around each of the existing training points, that is a total of \new{${N}_{epi} = n{N}_{tr}$} from a normal distribution 
such that:
\begin{align}
	\label{eq:def_sXepi}
	\siXepi = \bigcup\limits_{i=1}^{\Ntr} \left\{
	\xepi_j \sim \distN(\vxtr_i,\mathrm{diag}(\bm{\nu})), j=1,\ldots,n
	\right\},
\end{align}
where $\bm{\nu}$ is a parameter vector defining a diagonal covariance matrix.
Hyper-parameters $n$ and $\boldsymbol{\nu}$ must be judiciously chosen to ensure that we train $\EpiOut$ to be robust with respect to the unknown function $\vx$ we are attempting to model.

In the following, we lay out our proposals for the choice of these hyper-parameters and the intuitions behind them.

\subsection{Choice of $\boldsymbol{\nu}$, $n$}

$\boldsymbol{\nu}$ can be interpreted as an analogue to the lengthscale \new{vector} of a \new{squared exponential kernel in a} GP, that is, a measure for how far away from a training point the prediction is influenced: 
Larger~$\boldsymbol{\nu}$ will lead to further spread of~$\sXepi$ and, therefore, low epistemic uncertainty in the neighborhood of the training data.
A natural choice for $\boldsymbol{\nu}$ therefore can be the inverse of the gradient magnitude of the true mapping $\vf$. If the gradient of the true mapping $\vf$ at a point $\vx$ in the input domain were large in magnitude, it would indicate that the value of $\vf(\cdot)$ varies more quickly around the point $\vx$ as opposed to the case where this gradient were smaller.
While the true mapping $\vf$ is unknown, a model $\hat \vf$ trained on $\mathbb{D}_{tr}$ to approximate $\vf$ is accessible. We therefore propose to choose $\boldsymbol{\nu}$ to be a function of the input point $\vxtr_i \in \mathbb{X}_{tr}$ we sample around as well as the gradient of the trained model $\hat \vf$. 
$\boldsymbol{\nu}$ can therefore be selected as:


\begin{align}
	\label{eq:def_nu}
	\boldsymbol{\nu}(\vxtr_i, \hat \vf) = {\boldsymbol{1}^{d_x}}\oslash\left({ \begin{bmatrix} \|\nabla_{\bm{x}} \hat{f}_1({\bm{x}}_i)\|_{\infty}\\ \vdots\\ \|\nabla_{\bm{x}} \hat{f}_{d_x}({\bm{x}}_i)\|_{\infty} \end{bmatrix} + c \cdot \boldsymbol{1}^{d_x}}\right),
\end{align}

where $\boldsymbol{1}^{d_x}$ is a vector of size $d_x$ with all elements equal to $1$, $\oslash$ is the element-wise (Hadamard) division operator and $c$ is a positive constant whose choice bounds the maximum value of $\boldsymbol{\nu}$. It can generally be set to a small positive value $\ll1$ in order to avoid division-by-zero errors, in the event the gradient at a point is $0$. In an online learning setting, the value of $c$ can be chosen to be large~$\approx1$ when the number of training points is too low and $\nabla_\vxtr\hat \vf(\vxtr_i), \vxtr_i \in \mathbb{X}_{tr}$ cannot be reliably estimated. This value can then be annealed to a small value $c\ll1$ as the gradient estimate becomes more reliable. Finally, the uniform norm $||\cdot||_{\infty}$ selects the maximum element from a vector. Applying this to every column of the Jacobian matrix $\nabla_\vxtr\hat \vf$ yields a vector of $d_x$ dimensions. Thus if the magnitude of the model's slope at a training point $\vxtr$ is high for one of the output dimensions with respect to a particular input dimension, we sample points for $\sXepi$ closer to $\vxtr$.  




We propose setting~$n$ to a multiple of~$2\dx+1$. This corresponds  to padding each training point in both directions of the input dimension with an OOD point on an average for the generation of the epistemic dataset $\sDepi$. The next subsection discusses the final steps required for completing the creation of the epistemic dataset $\sDepi$. 

\subsection{Generation of $\mathbb{Y}_{epi}$}\label{subsec:yepi}
To define the set $\sYepi$, we first compute the minimal distance (according to a distance metric~$d\colon \sX \times \sX \to \sRzp$) to the training data for each epi point.
\begin{align}
	\label{eq:distance}
	d_j = \min_{\vxtr \in \sXtr} d(\xepi_j,\vx), \quad j = 1,\ldots,\Nepi.
\end{align}
Note that for an epi point $\tilde{x}_j$, the closest training data point is not necessarily the one used to generate it.
Let $d_\Ntr$ be the $\Ntr$-th smallest element of the set $\bm{d} = \{d_j\}_{j=1}^{\Nepi}$. We generate~$\sYepi$ and derive~$\sXepi$ from $\siXepi$ as follows
\begin{align}
	\label{eq:def_yepi}
	\yepi_j =  \left\{ \begin{array}{lll}
		1,  &   &\text{if } d_j > d_\Ntr \\
		0,  & \xepi_j \leftarrow \argmin_{\vxtr \in \sXtr} d(\xepi_j,\vx) 
		&\text{if 
		} d_j \leq d_\Ntr \\
	\end{array} \right. .
\end{align}
The $\Ntr$ points in~$\siXepi$ with the least distance to a training point are therefore replaced by the corresponding point in~$\sXtr$. This step attempts to reduce the overlap between the input space covered by $\mathbb{X}_{epi}\backslash\mathbb{X}_{tr}$ and the space covered by the training inputs $\mathbb{X}_{tr}$. Now the choice of~$2\dx+1$ epi points becomes clear as one of them will be turned into $\yepi = 0$ corresponding to ``low epistemic uncertainty'', while $2\dx$ points further away from the training point $\yepi = 0$ correspond to ``high epistemic uncertainty''. See Fig. \ref{fig:NN}(b) for an example of epistemic training data generation.

\subsection{Overall architecture and training}

Our uncertainty aware model for disturbance $\vy$ is implemented via a neural network with outputs~$[\hat{\vf}(\cdot)\ \hat{\bm{\sigma}}(\cdot) \ \epi(\cdot)]^T$, where $\hat{\vf}(\cdot)$ corresponds to the underlying noiseless true mapping $\vf$, $\hat{\bm{\sigma}}(\cdot)$ is the aleatoric uncertainty estimate, and, $\epi(\cdot)$ corresponds to the epistemic uncertainty estimate (see Fig.~\ref{fig:NN} for an illustration). The new output $\epi(\cdot)$ is terminated with a neuron using a sigmoidal activation function, such that $\epi\colon\sX \to [0, 1]$. This is beneficial because it allows one to immediately judge whether the predicted uncertainty is \textit{high} ($\approx 1$) or \textit{low} ($\approx 0$) without any reference which would be required for unnormalized uncertainty quantifications.

We first train the feature layers as well as the output and the aleatoric uncertainty layers ($\hat \vf, \hat{\bm{\sigma}}$) by minimizing a negative log likelihood loss with respect to training data $\mathbb{D}_{tr}$, similar to \citet{kendall2017what}.
The gradient of the trained network can then be utilized to calculate $\boldsymbol{\nu}$ and sample the dataset $\mathbb{D}_{epi}$ in accordance with the method laid out in the previous subsections.

The epistemic module ($\epi$) is trained independently of the loss function for the original network over the dataset $\mathbb{D}_{epi}$ using a binary cross-entropy loss, which is the natural choice for binary outputs. It quantifies the uncertainty in the prediction of the other outputs based on the distance to the training data measured by~$d(\cdot,\cdot)$. For the sake of simplicity, we use the Euclidean distance, however the method can be easily extended to other metrics. Since the number of epistemic data-points with uncertain labels (with value $1$) are far greater than the ones with certain labels (with value $0$), a class-weighted version of this loss is minimized.

\subsection{Computational complexity}
\label{sec:compute_complex}
Generation of the pseudo data-set \eqref{eq:def_sXepi} is an operation with complexity $\bigO(\Nepi)\widehat{=}\bigO(\Ntr\dx)$, whereas~\eqref{eq:distance} is, for a trivial implementation,
a $\bigO(\Ntr\Nepi)\widehat{=}\bigO(\dx\Ntr^2)$ operation. However, a kd-tree based implementation~\citep{cormen2009introduction} allows for an 
execution time complexity~$\bigO(\Nepi\log(\Nepi))\widehat{=}\bigO(\dx\Ntr\log(\Ntr\dx)))$. Finding the~$\Ntr$ smallest distances from all~$\Nepi$ points in \eqref{eq:def_yepi} can be obtained in~$\bigO(\Ntr + (\Nepi-\Ntr)\log(\Ntr))\\\widehat{=}\bigO(\Ntr + \Ntr(\dx-1)\log(Ntr))$ time. The training of a neural network with a fixed number of weights requires~$\bigO(\Nepi)\widehat{=}\bigO(\Ntr\dx)$.
Hence, the overall complexity results in~$\bigO(\dx\Ntr\log(\dx\Ntr))$, and the overall storage complexity is given by~$\bigO(\Nepi\dx)\widehat{=}\bigO(\Ntr\dx^2)$ for storing the set~$\sXepi$. 
When considering streaming data (as for online learning control), the set $\sDepi$ can be constructed iteratively, reducing the complexity to~$\bigO(\log(\Ntr))$

\section{Real-time Control with Decomposed Uncertainties}
\label{sec:online_control}

We can now deploy the proposed approach for uncertainty decomposition in a real-time control loop. We utilize the \EpiOut~module and its output $\epi$ for adding new data points for combating epistemic uncertainty in a data-efficient manner, while exploiting the aleatoric uncertainty estimates $\hat{\bm{\sigma}}$ for adjusting the controller gain to measurement noise. The latter robustifies the closed-loop control mechanism against process noise and can even guarantee stability~\citep{beckers2019stable}, while  keeping the energy consumption low. The former combats epistemic uncertainty via increased data collection. 


Considering the dynamical system given in \eqref{eq:sys}, our task is to choose a control input~$\vu$ at each step, such that the system follows a given reference $\vx^{\mathrm{des}}$. 
Furthermore, the controller can take a new measurement $\vy$ and add $\{\vx, \vy\}$ to the training dataset to improve its
model over time but each such measurement is considered costly and therefore new training points should only be collected when necessary. This is typical in distributed systems, where multiple sensors share the same scarce communication channel, or in autonomous systems with limited data storage capacity.

\subsection{Epistemic uncertainty utilization for data-efficient model learning}

The need for high data efficiency requires models that can judge their own 
fidelity in real-time to identify valuable measurements.
We utilize \EpiOut~by adding new measurements~$(\vx_i,\vy_i)$ to the training dataset based on the epistemic uncertainty $\epi(\cdot)\in[0,1]$ of the data-point. More concretely, we use the epistemic uncertainty to parametrize a Bernoulli distribution which is used to determine whether to add the measurement to the training dataset or not, 
\begin{equation}
	\begin{split}
		\label{eq:accept_xtr}
		\sDtr \leftarrow \sDtr \cup 
		\begin{cases}
			(\vx_\ri, \vy_\ri) & \text{if } \xi = 1 \\
			\emptyset & \text{if } \xi = 0
		\end{cases},
		\\\text{where } \xi \sim \distB(\acp), \acp= \epi(\vx_\ri).
	\end{split}
\end{equation}

This ensures the high accuracy of the disturbance model $\hat{\vf}$ while at the same time bolstering data-efficiency in the online data collection process, as training data is added only when necessary, thus reducing the number of costly measurements. 

\subsection{Aleatoric uncertainty based gain adjustment}

The system \eqref{eq:sys} is inherently random due to the stochastic nature of
the disturbance $\vy$. Therefore, we combine feedback and a feedforward control law
\begin{align}
	\label{eq:ctrl}
	\vu = \bm{K}\left(\vx-\vx^{\mathrm{des}}\right) + \vu_{\text{ff}},
\end{align}
where 
$\vu_{\text{ff}}$ is a feedforward control term determined based on the known 
model 
$\vg(\cdot,\cdot)$ and the learned disturbance model~$\hat{\vf}(\cdot)$.
The choice of the feedback gain~$\bm{K}$, which compensates for imprecision in the model and the stochasticity of the disturbance, is difficult because high control gains lead to high tracking performance, but also consume more energy and amplify measurement noise, which can lead to increased tracking errors. It is therefore, generally advisable to let the feedforward term~$\vu_{\text{ff}}$ take over most of the control effort and keep 
the feedback term small when the model is reliable (see \citet{beckers2019stable}). We thus increase the 
gains only if the aleatoric uncertainty inferred by our model as~${\hat{\bm{\sigma}}}(\cdot)$ is high. The feedback gains can therefore be adjusted as:
\begin{align}
	\label{eq:vargain}
	\bm{K} = \bar{\bm{K}}\big(1 +\beta||\hat{\bm{\sigma}}(\vx)||_\infty),
\end{align}
where~$\beta \in \sRzp$ is the sensitivity and $\bar{\bm{K}}\in\mathbb{R}^{d_u\times d_x}_+$ defines the minimal control gain. 


Therefore, we tune the feedback gains only based on the aleatoric uncertainty, while combating the epistemic uncertainty with an increased data collection rate~\eqref{eq:accept_xtr}.

\section{Evaluation}
\label{sec:Eval}

We evaluate our proposed approach for estimating epistemic uncertainty both qualitatively and quantitatively for a variety of synthetic and real datasets and compare them to existing approaches. We then demonstrate the utility of our decomposable uncertainty estimation approach for data-efficient online learning and control on a simulated quadcopter system affected by unknown disturbances.

\subsection{Qualitative and Quantitative Evaluation for Regression}
\label{sec:compare_exps}

Our approach is evaluated on two simulated and one real-world dataset for illustrating the advantage of our approach when estimating uncertainty due to absence of data. Both are 1-dimenstional regression tasks. The first, \textit{1D Split}, has the nominal function $f(x) = \sin(x\pi)$, with training points located only in 2 bands $\sXtr  = \{\xtr_i \sim 	\distU(-2,-1)\}_{i=1}^{100}\cup \{\xtr_i \sim \distU(1,2)\}_{i=101}^{200}$ and~$\Nte=961$ test points are placed on a grid~$[-4, 4]$. The second, \textit{2D Gaussian} ($\dx=2$, $\dimp=1$) has the nominal function  $f(\vx) = \frac{\sin(5x_1)}{5x_1} + x_2^2$.  Two clusters of $500$ points each are respectively sampled from Gaussian distributions placed around points $(0, -1)$ and $(0, 1)$. $961$ test points are uniformly placed on a grid~$[-2, 2]^2$. We additionally evaluate our approach on a real dataset \textit{Sarcos} \citep{vijayakumar2000locally} which records the inverse dynamics of a seven degrees-of-freedom SARCOS anthropomorphic robot arm $\dx= 21$, $\dimp= 1$. $10000$ training samples and $2000$ test samples were randomly extracted  from a  total of~$5\times 10^4$ samples.  

We compare our approach to several state-of-the-art methods. These include a Gaussian Process (\textit{GPmodel}) with a squared exponential automatic relevance determination kernel, a Bayesian Neural Network (\BNN) with $2$ fully connected hidden layers each with $50$ hidden units and normal distributions over their weights, a neural network with $2$ fully connected permanent layers each with $50$ hidden units with dropout probability~$\rho= 0.05$ (\Dropout). Our proposed  model \EpiOut~has $2$ fully connected layers ($50$ neurons) ,~$n= 3$ and $\boldsymbol{\nu}$ as given by \eqref{eq:def_nu}, where the value $c$ is set to \new{$10^{-5}$}. 

\begin{figure}
	\centering 
    \subfigure[\small 1D Split Dataset]{%
  \def\FILENAME{1D_split.dat}%
  \def\YMIN{-0.1}%
  \def\YMAX{1.1}%
  \begin{tikzpicture}
\begin{axis}[
    ymin = \YMIN, ymax = \YMAX,
    xtick distance = 5,
    table/col sep=semicolon,
    grid = both,
    minor tick num = 1,
    major grid style = {lightgray},
    minor grid style = {lightgray!25},
    width = 0.45\textwidth,
    height = 0.25\textwidth,
    legend cell align = {left},
    legend pos = north west,
    label style={scale=0.8, text width=2cm},
    ylabel={Epistemic Uncertainty},
legend style={nodes={scale=0.6},at={(0.5,1.2)}, anchor=north,legend columns=-1,
                /tikz/every even column/.append style={column sep=0.3cm}},
]
 
\addplot[nicegreen, ultra thick, dashed] table [x = {GPmodel_x}, y = {GPmodel_epi}] {data/\FILENAME};
\addlegendentry{GPmodel}
\addplot[orange, very thick, dotted] table [x = {BNN_x}, y = {BNN_epi}] {data/\FILENAME};
\addlegendentry{BNN}
\addplot[violet, semithick, solid] table [x = {Dropout_x}, y = {Dropout_epi}] {data/\FILENAME};
\addlegendentry{Dropout}
\addplot[teal, ultra thick] table [x = {EpiOut_x}, y = {EpiOut_epi}] {data/\FILENAME};
\addlegendentry{EpiOut}
\addplot[only marks,scatter,color=blue] table [x = {xtr}, y = {xtr_y}] {data/\FILENAME};
\addlegendentry{$\mathbb{X}_{tr}$}
\end{axis}
\end{tikzpicture}%
}\\
        \subfigure[\small 2D Gaussian Dataset]{\includegraphics[width=0.47\textwidth, trim={0 4cm 0 5cm},clip]{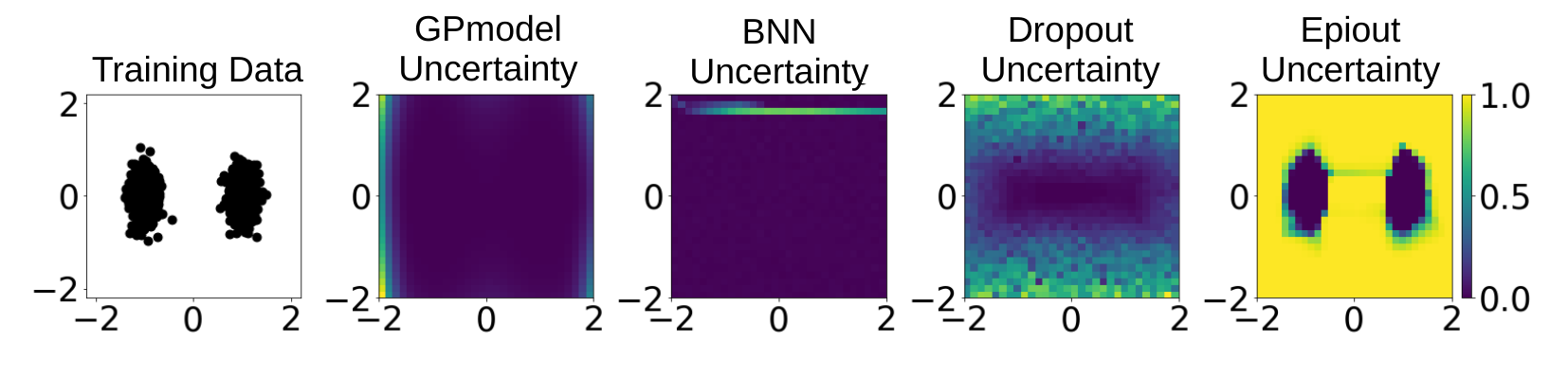}}	
	\caption{The epistemic uncertainty given by different models for datasets (a) 1D Split and (b) 2D Gaussian. Note that unlike EpiOut, other methods are unable to predict a higher uncertainty between the training data clusters.} 
	\label{fig:1D_split}
\end{figure}

\begin{figure}
	\centering
    \begin{tikzpicture}
\begin{axis}[
  legend pos=north west,
  xtick=data,
  table/col sep=semicolon,
  ybar,
  bar width=0.25cm,
  legend style={nodes={scale=0.7},at={(0.5,1.4)}, anchor=north,legend columns=-1,
                    /tikz/every even column/.append style={column sep=0.3cm}},
  width = 0.42\textwidth,
  height = 0.2\textwidth,
  enlarge x limits={abs=1500},
  ylabel={Prediction Time (s)},
  xlabel={Number of Training Data Points},
  ylabel near ticks,
  label style={scale=0.8},
  ticklabel style = {font=\tiny},
  error bars/y dir=both, 
  error bars/y explicit  
  ]

  \addplot[nicegreen!20!black,fill=nicegreen!80!white, postaction={
    }] table[x=data_sz,y=GPmodel] {data/sarcos_time.dat};
  \addlegendentry{GPmodel}

  \addplot[orange!20!black,fill=orange!80!white, postaction={
    }] table[x=data_sz,y=BNN] {data/sarcos_time.dat};
  \addlegendentry{BNN}

  \addplot[violet!20!black,fill=violet!80!white, postaction={
    }] table[x=data_sz,y=Dropout] {data/sarcos_time.dat};
  \addlegendentry{Dropout}
  
  \addplot[teal!20!black,fill=teal!80!white] table[x=data_sz,y=EpiOut] {data/sarcos_time.dat};
  \addlegendentry{EpiOut}

\end{axis}
\end{tikzpicture}
	\caption{The prediction time for various models with increasing training data (Sarcos). As is evident, \EpiOut~outperforms all other models in this regard.}
	\label{fig:tevaluate}
\end{figure}

\if \tblincmse0
    \begin{table}
	\caption{Mean Standardized Log loss for the considered models across datasets.
	}
	\label{tab:msll}
	\begin{center}
				\begin{tabular}{lVrVrVr}
					\toprule
					{} &  1D Center& 1D Split & 2D Square & 2D Gaussian & PMSM & Sarcos \\
					\midrule
					GPmodel &     {-3.8075} 
					&  \textbf{-3.69} 
					&  {-2.6602} 
					&  {873.21} 
					&  {-2.4028} 
					&  {0.1534} \\
					BNN     &     0.57 
					&  {0.73} 
					&  \textbf{-0.60} 
					&  -75.6
					&  -1.80 
					&  -1.37\\
					Dropout &     5.27
					&  2603
					&  653
					&  2.34 
					&  \textbf{-1.87} 
					&  -1.65 \\
					EpiOut  &     \textbf{-0.01} 
					&  {-0.08} 
					&  {2.34} 
					&  \textbf{-76.7} 
					&  {4.36} 
					&  \textbf{-1.74} \\
					\bottomrule
				\end{tabular}
	\end{center}
\end{table}

\else
    \begin{table}
	\caption{Evaluation of considered models with MSLL and MSE across datasets.}
	\label{tab:msll}
	\begin{center}
		\begin{tabular}{lcccccc}
		\toprule
        \multicolumn{1}{c}{} & \multicolumn{2}{c}{ Split } & \multicolumn{2}{c}{ Gaussian } & \multicolumn{2}{c}{ Sarcos } \\
        \multicolumn{1}{c}{} & MSLL & MSE & MSLL & MSE & MSLL & MSE \\
        \midrule
        GPmodel  & -3.69 & 0.46 & 873.21 & 0.03 & 0.15 & 5.1 \\
        BNN  & 0.74 & 0.74 & -75.62 & 4.87 & -1.37 & 28.93 \\
        Dropout  & 2603.38 & 1.53 & 652.56 & 0.68 & -1.66 & 12.73 \\
        EpiOut  &-0.08 & 0.89 & -76.77 & 0.39 & -1.74 & 9.21 \\
        \bottomrule
		\end{tabular}
	\end{center}
\end{table}

\fi

\subsection{Results \& Discussion}

Evaluation of the proposed method using Mean Standardized Log Likelihood (MSLL)~\citep{rasmussen2006gaussian}
\if \tblincmse1
and Mean Squared Error (MSE)
\fi
is reported in Table \ref{tab:msll}. 
Since our approach predicts values for uncertainty constrained between $[0,1]$, for fair comparison we calculated  MSLL for \EpiOut, by scaling the uncertainty estimate by a scalar value that is optimized to maximize the log-likelihood of the training data. We note that \EpiOut~demonstrates competitive performance when evaluated using MSLL for most of the datasets. 
\if \tblincmse1
MSE is only indicative of the predictive error of the point estimate.
\fi

 Additionally, an illustration of epistemic uncertainty for datasets \textit{1D Split} and \textit{2D Gaussian} for all models is shown in Fig.~\ref{fig:1D_split}. Note that, in contrast to other approaches, \EpiOut~is more successful in identifying out of distribution data.
 We further point out additional benefits of our approach: (i) The \EpiOut~model predicts the uncertainty measure in a sample-free manner and as a result, in real time. This is typically an order of magnitude faster than \Dropout~and \BNN~(see Fig.~\ref{fig:tevaluate}). This is crucial for data-efficient online learning, where the epistemic uncertainty estimate is used to evaluate the usefulness of an incoming data point and is thus calculated more frequently than the online training is executed. (ii) A single evaluation of~$\epi(\cdot)$ is sufficient for  a conclusion whether the uncertainty is high or low, since it is bounded to the interval~$[0,1]$, whereas alternative approaches provide a return value~$[0,\infty]$, which can be difficult to interpret without a maximum value as reference.



\subsection{Event-triggered Online Learning Control Using \\Uncertainty Decomposition}
\label{sec:quadcopter_control}

As an application of our proposed approach, we consider the task of controlling a quadcopter with known dynamics $\vg(\cdot, \cdot)$ \eqref{eq:sys}, which explores a given terrain with unknown thermals $\vy$\footnote{The data of the thermals is taken from publicly available paragliding data https://thermal.kk7.ch. Thermals or thermal columns are created by the uneven heating of the earth's surface and cause a vertical air flow due to convection~\citep{akos2010thermal}.}$\!$. Thus $\vy$ is an indicator of the thermal disturbance on the quadcopter acceleration in vertical direction. We assume that it follows the underlying distribution \eqref{eq:true}.


The task here is to execute the desired trajectory~$\vx^{\mathrm{des}}$ which we specify as three rounds at constant height on a square in the horizontal plane with edge length~$0.1$ followed by three rounds with edge length~$0.05$.

Following the method laid out in section \ref{sec:online_control} for implementing a data-efficient control strategy, the tracking performance of the proposed quadcopter controller, the data collections rate and the epistemic uncertainty model are illustrated in Fig.~\ref{fig:xxdes}. 
The \EpiOut~prediction of the disturbance model allows smart selection of the measurements taken for training. While a uniformly random selection constantly takes data points, as shown in Fig.~\ref{fig:xxdes}(a), our proposed approach takes more measurements when entering new areas ($0s$ to $5s$ and $12s$ to $17s$) and less when tracking the path already traversed. This is in accordance with the epistemic uncertainty estimate $\eta$ which is low near training data and high in unobserved regions as is observable in Fig.~\ref{fig:xxdes}(c). As illustrated in Fig.~\ref{fig:xxdes}(b), this results in overall competitive tracking performance of a root mean squared error in z-direction of 0.0079 vs 0.0086 while using and storing less data.

\if\loadsomeimgs1

\begin{figure}
	\centering 
	\subfigure[\small{Number of training datapoints acquired when sampling using  \EpiOut~vs via a uniform distribution}]{\includegraphics[width=0.46\textwidth]{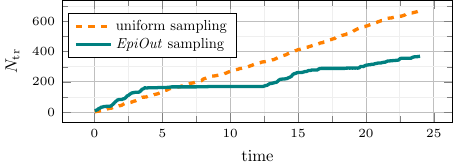}}
	\subfigure[\small Root Mean Square Error during tracking]{\includegraphics[width=0.16\textwidth]{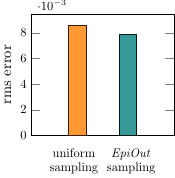}}
	\hspace{2mm}
	\subfigure[\small{Epistemic uncertainty estimate ($\eta \in [0,1]$) over the horizontal plane}]{\includegraphics[width=0.25\textwidth]{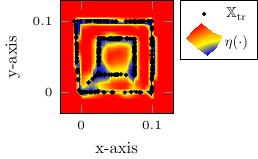}}
	
	\caption{Deploying the control strategy from section \ref{sec:online_control} with \EpiOut~results in data-efficiency (a) while demonstrating competitive performance to the method which utilizes all available data for training (b). Epistemic uncertainty over the horizontal plane, indicated by absence of training data and estimated at the end of the task is shown in (c).}
	\label{fig:xxdes}
\end{figure}
\else

\begin{figure}
	\centering
	\centering

\def\fileA{data/quad_noepi.dat}
\def\fileB{data/quad_epi_non_uni.dat}
\def\fileC{data/dmodel_non_uni.csv}
\subfigure[\small{Number of training datapoints acquired when sampling based on  \EpiOut~vs a uniform distribution}]{
\centering
\tikzsetnextfilename{quad_numdata}
\begin{tikzpicture}
	
	\begin{axis}[
        table/col sep=semicolon,
		xlabel={time},
		ylabel={$\Ntr$},
        xtick distance = 5,
        grid = both,
        ylabel near ticks,
        xlabel near ticks,
        label style={scale=0.8},
        minor tick num = 1,
        major grid style = {lightgray},
        minor grid style = {lightgray!25},
        width = 0.45\textwidth,
        height = 0.2\textwidth,
        legend cell align = {left},
        legend pos = north west,
        ticklabel style = {font=\tiny},
  legend style={nodes={scale=0.7},at={(0.23,0.9)}, anchor=north,legend columns=1,
                    /tikz/every even column/.append style={column sep=1.0cm}},
		]
		\addplot[orange, ultra thick, dashed] table [x=t, y=ntr] {\fileA};
		\addplot[teal, ultra thick] table [x=t, y=ntr] {\fileB};
		\legend{uniform sampling, \EpiOut~sampling}
	\end{axis}
	
\end{tikzpicture}%
}

\subfigure[\small Root Mean Square Error during tracking]{
\centering
\tikzsetnextfilename{quad_rmse}
\begin{tikzpicture}
\begin{axis}[
  legend pos=north west,
ylabel near ticks,
xlabel near ticks,
  xmin=0,
  xmax=1,
  ymin=0,
  ybar,
  xtick={0.3, 0.7},
  x tick label style={align=center, text width=1.5cm},
  x tick style={draw=none},
  xticklabels={uniform sampling, \EpiOut~\\sampling},
  bar width=0.3cm,
  width = 0.22\textwidth,
  height = 0.2\textwidth,
  ylabel={rms error},
  ylabel near ticks,
  label style={scale=0.7},
  ticklabel style = {scale=0.6},
  error bars/y dir=both, 
  error bars/y explicit  
  ]

\addplot[orange!20!black,fill=orange!80!white] coordinates {(0.4,0.0086)};
\addplot[teal!20!black,fill=teal!80!white] coordinates{(0.6,0.0079)};

\end{axis}
\end{tikzpicture}
}

\hfill

\subfigure[\small{Epistemic uncertainty estimate ($\eta \in [0,1]$) over the horizontal plane}]{
\centering
\tikzsetnextfilename{quad_epi}
\begin{tikzpicture}
    \begin{axis}[
        xlabel={x-axis}, ylabel={y-axis}, 
        ylabel near ticks,
        xlabel near ticks,
        label style={scale=0.8},
        title style={scale=0.8},
        width = 3.5cm,height =3.5cm,
        legend style={nodes={scale=0.7},at={(1.4,1.0)}, anchor=north,legend columns=1,
                    /tikz/every even column/.append style={column sep=1.0cm}},
        ticklabel style = {font=\tiny},
        grid = none, view={0}{90},
        xmin = -0.03, ymin = -0.03, zmin = 0, 
        xmax = 0.13, ymax=0.13, zmax=1.2,
        ]
        \addplot3[only marks,black, mark size=0.8pt] table 
        		[x=xtr_0, y=xtr_1, z = ytr_0] {\fileC};
        \addplot3[surf,shader=interp,mesh/rows=60, opacity = 1] table 
        [x=xte_0,y=xte_1,z= epi_modelte]{\fileC};
        \legend{$\sXtr$,$\epi(\cdot)$}
    \end{axis}
\end{tikzpicture}
}

\caption{Deploying the control strategy from section \ref{sec:online_control} with \EpiOut~results in data-efficiency (a) while demonstrating competitive performance to the method which utilizes all available data for training (b). Epistemic uncertainty over the horizontal plane, indicated by absence of training data and estimated at the end of the task is shown in (c).} 
\label{fig:xxdes}
	\label{fig:xxdes}
\end{figure}
\fi

\section{Conclusion}
\label{sec:conc}
This paper presents a novel deep learning structure for decomposing epistemic 
and aleatoric uncertainty and proposes a control framework that distinguishes between the use of  these uncertainty measures. As the predictions and uncertainty estimates are obtained in a sample-free manner, the method can be utilized for real-time critical online learning and shows competitive performance against existing methods. The proposed online learning control algorithm is not only inherently data-efficient by adding only those data-points to the data set that can reduce the model's epistemic uncertainty, but is additionally able to provide a means to adapt the control gains for better tracking performance by taking the aleatoric uncertainty estimates into consideration.

For future work, we will consider alternative functions for sorting the epi points to encode prior knowledge, such as periodicity (similar to a kernel of GP) and investigate the effect of a continuous-valued epistemic uncertainty label~$\mathbb{Y}_{epi}$. Additionally, we will explore the application of our uncertainty estimation approach for different contexts, for example in scenarios where instead of unknown system dynamics, other aspects such as part of the cost or system constraints are unknown.

\bibliography{ifacconf}             
                                                   







\end{document}